\documentclass[a4paper]{svproc}
\usepackage{style}
\usepackage{url}

\usepackage{algorithm2e}

\begin{document}
\mainmatter              
\title{Learning a Stable, Safe, Distributed Feedback Controller for a Heterogeneous Platoon of Autonomous Vehicles}
\titlerunning{Learning Platoon Distributed Control}  
%
\author{Michael H. Shaham \and Ta\c{s}k{\i}n Pad{\i}r}
\authorrunning{Michael H. Shaham and Ta\c{s}k{\i}n Pad{\i}r} 
%
\tocauthor{Michael H. Shaham and Ta\c{s}k{\i}n Pad{\i}r}
\institute{Northeastern University, Boston, MA 02116, USA\\
\email{shaham.m@northeastern.edu}\\ 
\texttt{https://robot.neu.edu/}
}

\maketitle              

\begin{abstract}
Platooning of autonomous vehicles has the potential to increase safety and fuel efficiency on highways. The goal of platooning is to have each vehicle drive at a specified speed (set by the leader) while maintaining a safe distance from its neighbors. Many prior works have analyzed various controllers for platooning, most commonly linear feedback and distributed model predictive controllers. In this work, we introduce an algorithm for learning a stable, safe, distributed controller for a heterogeneous platoon. Our algorithm relies on recent developments in learning neural network stability certificates. We train a controller for autonomous platooning in simulation and evaluate its performance on hardware with a platoon of four F1Tenth vehicles. We then perform further analysis in simulation with a platoon of 100 vehicles. Experimental results demonstrate the practicality of the algorithm and the learned controller by comparing the performance of the neural network controller to linear feedback and distributed model predictive controllers.

\end{abstract}

\section{Introduction}

In safety-critical systems where performance guarantees are of paramount importance, adoption of learning-based controllers has understandably been slow. This is because most learning-based controllers, up until recently, have not had the ability to provide performance guarantees over the range of scenarios the system can expect to encounter. Recent work in learning controllers, safety certificates, and stability certificates has opened the door for learning reliable controllers for safety-critical systems~\cite{dawson2023learncertificates}. As we consider using these controllers in real systems, however, developing algorithms that will learn controllers that translate well from simulation to the real world is critical. Furthermore, for these methods to work on multi-agent systems, it will be necessary to develop algorithms that scale as the number of agents increases.

Autonomous vehicle platooning is an emerging driving technology that can improve safety~\cite{magdici2017accsafety} and reduce fuel emissions~\cite{liang2016platoonfuel} on our roads. However, there is an inherent tradeoff between these two objectives as fuel efficiency is maximized when vehicles are closer together while safety is maximized when vehicles are further apart. This tradeoff between performance and safety makes the platooning problem a good case study for certifiable learning-based control. 

In this work, we focus on learning decentralized feedback controllers for a heterogeneous (\ie, each vehicle can have different dynamics) platoon of vehicles. The focus is on developing controllers that are provably stable for the entire platoon (learn a centralized stability certificate based on a decentralized controller). The difficulty in learning a stable decentralized controller for a multi-agent system where we need centralized performance guarantees arises from the fact that the computational complexity of verifying stability grows exponentially as the size of the system (number of vehicles) grows. This is because verifying stability requires solving mixed-integer linear programs (MILPs).

\begin{figure}
    \centering
    \includegraphics[width=0.60\textwidth]{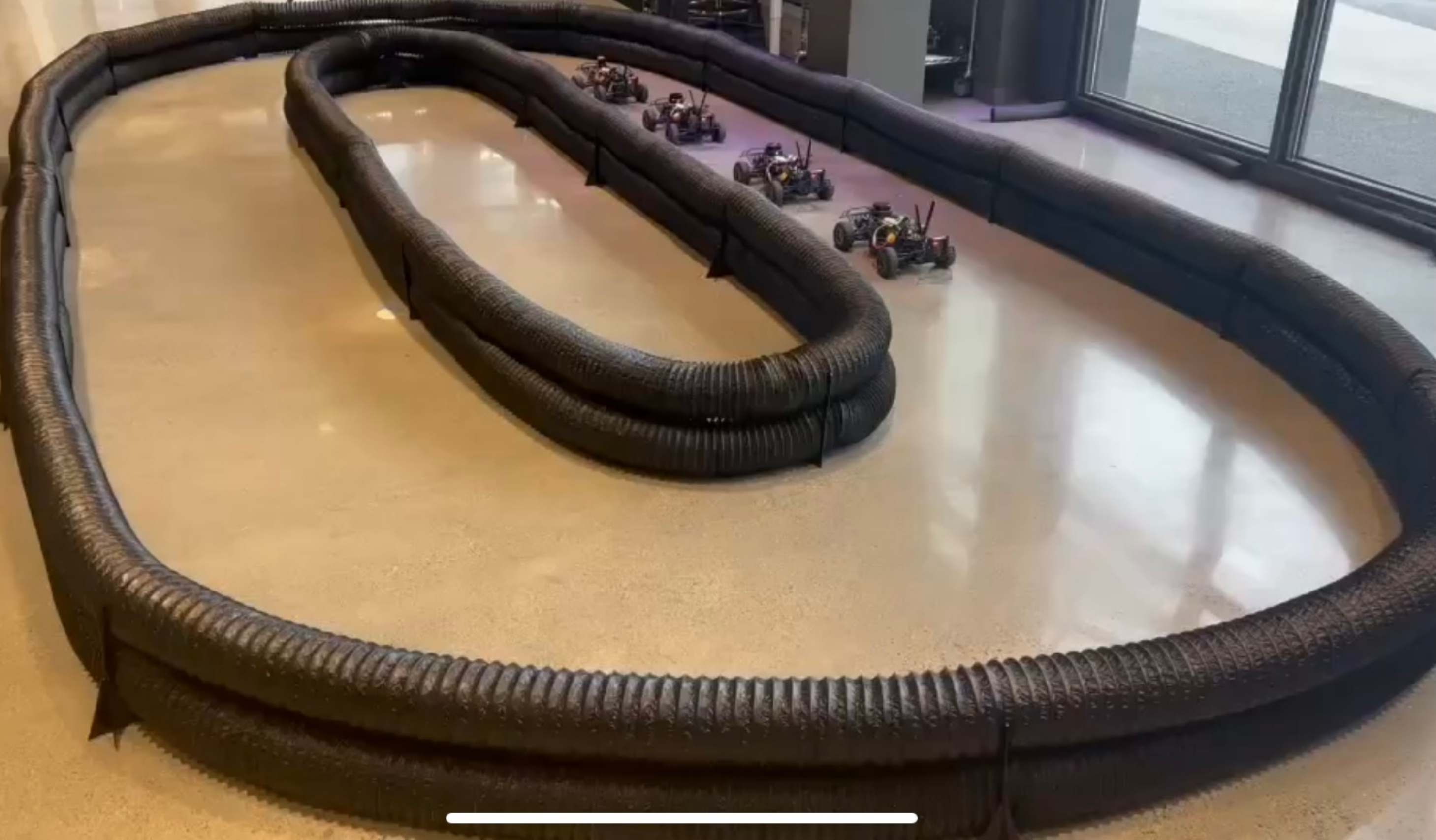}
    \caption{The platoon of four F1Tenth vehicles in the test course.}
    \label{fig:convoy_exp}
    \vspace{-6.0mm}
\end{figure}

The contributions of this work are as follows: 1) we introduce an algorithm for learning a certifiably stable neural network controller that enables the engineer to ``guide'' the resulting controller to desired behavior; 2) we use a change of variable to reformulate the dynamics of a heterogeneous platoon as that of a homogeneous platoon, enabling us to learn and verify a single controller for a heterogeneous platoon; 3) we train controllers in simulation and validate the learned controllers on real hardware using four F1Tenth vehicles~\cite{okelly2020f1tenth}, shown in~\cref{fig:convoy_exp}, and in simulation with 100 vehicles.

\section{Related Work}

\subsection{Autonomous vehicle platooning}

The platooning problem first began gaining attention from a theoretical perspective as early as the 1960s when the authors of~\cite{levine1966errorregulation} analyzed the performance of a 3-vehicle platoon using a constant distance headway (CDH) and a linear feedback controller. However, it wasn't until the 1990s and the California PATH demonstration~\cite{tan1997pathdemo} when the problem began gaining widespread attention. Much of the prior work has focused on gaining theoretical insights that guide algorithm development. 

In~\cite{seiler2004distprop}, the authors showed that using a linear feedback controller and a CDH spacing policy can lead to string instability, meaning disturbances witnessed earlier in the platoon will propagate and become worse toward the end of the platoon. Additionally, they showed how the disturbance to error gain increases as the size of the platoon increases for both a predecessor following (PF) and bidirectional (BD) topology. Building on this, the authors of~\cite{zheng2016stabilityhomogeneous} showed how the stability margins of a platoon using a linear feedback controller with BD topology decays to zero as $O(1/N^2)$ when each vehicle has the same dynamics. This work extended~\cite{hao2013stability}, where the authors also showed how a particular nonlinear controller outperformed linear controllers for a platoon using a PF topology.

Since CDH spacing policies lead to string instability, researchers switched their focus to constant time headway (CTH) spacing policies where the distance between vehicles is selected based on each vehicle's velocity. In~\cite{naus2010caccstringstable}, it was shown that platoons using linear feedback controllers with CTH spacing policies are string stable. However, it is common to use a time headway of at least 1.8 seconds~\cite{vogel2003comparison} which leads to large inter-vehicle distances and thus reduced traffic capacity~\cite{swaroop1994comparison}.

If vehicles can communicate with one another, distributed model predictive control (DMPC), which requires vehicles share their planned trajectories with their neighbors, is possible. One of the first DMPC controllers was analyzed in~\cite{dunbar2012distributedrhc} where it was shown that a DMPC controller is both stable and string stable for a PF topology with a CDH policy when all of the vehicles in the platoon know the desired velocity of the platoon a priori. In~\cite{zheng2017dmpc}, the stability condition was extended for the case that not all vehicles know the desired velocity a priori. Additionally, the work of~\cite{shaham2024dmpc} extended that of~\cite{zheng2017dmpc} for the case of CTH topologies and arbitrary communication topologies. In~\cite{shaham2024design}, two DMPC controllers, both slight deviations from the controller proposed in~\cite{zheng2017dmpc}, were validated on both hardware and in simulation. The simulation results, which analyzed the performance of DMPC as the size of the platoon scaled up to 100 vehicles, showed that DMPC greatly outperforms linear feedback controllers.

\subsection{Certifiable learning-based control}

Upon analyzing much of the prior theoretical results in platoon control, we arrive at the question: can we learn better controllers for platooning? Specifically, since many vehicles on the road today are equipped with forward-facing radar and adaptive cruise control technology, it would be beneficial to learn controllers for CDH spacing policies and PF topologies that close the performance gap between linear feedback and DMPC, thus enabling tightly-packed platoons and greater traffic capacity. To achieve this goal, we need to ensure these learned controllers are stable, which is where the emerging research areas of learning-based control~\cite{brunke2022safelearning} and neural network verification~\cite{liu2021nnver,everett2021nnvercontrol} merge into certifiable learning-based control. To date, research within certifiable learning-based control has generally been related to either reachability analysis~\cite{everett2021reachabilityanalysis} or learning Lyapunov and barrier certificates~\cite{dawson2023learncertificates}.

In this work, we are interested in learning the parameters of a neural network Lyapunov function to guarantee stability of a platoon using a distributed neural network controller. One of the earlier works in certifiable learning-based control was~\cite{chang2019neural}, where both a neural Lyapunov function and a controller were learned to prove stability when controlling a given nonlinear system. In~\cite{dai2021lyapstablenn}, this was extended to consider also systems with learned neural network dynamics. Depending on the system being analyzed, it can be straightforward to extend these earlier works directly to multi-agent systems, but the certification problem quickly becomes intractable as the number of agents (and thus the state size) increases due to the need to solve a nonconvex mixed integer program. In~\cite{qin2020learning}, the authors get around this due to the fact that the goal of each agent is independent of other agents, and thus stability certificates can be learned in a completely decentralized manner.

\section{Methodology}

\subsection{Controller design}

We consider a platoon of $N$ vehicles indexed by $i=1, \ldots, N$ where vehicle 1 is the leader and $i < j$ implies vehicle $i$ precedes vehicle $j$ in the platoon. Each vehicle has dynamics given by
\begin{equation}
    \label{eq:og_dynamics}
    s_i(k+1) = \bmat{1 & \dt \\ 0 & 1 - \frac{\dt}{\tau_i}} s_i(k) + \bmat{0 \\ \frac{\dt}{\tau_i}} v^\text{des}_i(k)
\end{equation}
where $s_i(k) = (p_i(k), v_i(k)) \in \R^2$ and $v^\text{des}_i(k) \in \R$ are vehicle $i$'s state (position and velocity) and control input (desired velocity), respectively, $\tau_i$ is a vehicle longitudinal dynamics delay parameter, and $\dt$ is the discrete timestep. It is common in the literature to use a three-state linear dynamics model with position, velocity, and acceleration as the state and desired acceleration as the control input, similar to~\cite{stankovic2000decentralized,zheng2016stabilityhomogeneous,zheng2017dmpc}. However, we use~\cref{eq:og_dynamics} for two reasons: 1)~it models the F1Tenth vehicle's longitudinal dynamics well and 2) it uses desired velocity instead of desired acceleration as the control input, similar to the F1Tenth vehicles.

We assume each vehicle in the platoon can sense the relative position and velocity of the vehicle in front of it. Our goal is to learn a distributed control law $\pi: \R^2 \to \R$ such that each vehicle $i$ drives at the same speed as its predecessor while maintaining some predefined distance $d_{i, i-1}$. The controller is distributed in the sense that each vehicle runs its controller locally based on information it receives from its sensors. The input to the controller is the error state 
\begin{equation}
\label{eq:error_state}
    x_i = \bmat{p_{i-1} - p_i - d_{i, i-1} \\ v_{i-1} - v_i}.
\end{equation}

We need to ensure each vehicle is able to use the learned controller in the same way even if individual vehicle dynamics are different due to the $\tau_i$ parameters. To do this, we define
\begin{equation}
    \label{eq:new_control}
    u_i(k) = \frac{v^\text{des}_i(k) - v_i(k)}{\tau_i}, \quad i = 1, \ldots, N.
\end{equation}
With this change of variable, the dynamics for vehicle $i$ become discrete-time double integrator dynamics given by
\begin{gather}
    \label{eq:new_dynamics}
    s_i(k+1) = A s_i(k) + B u_i(k),\\ 
    A = \bmat{1 & \dt \\ 0 & 1} \quad B = \bmat{0 \\ \dt} \notag
\end{gather}
where the control input $u_i \in \R$ is the acceleration. We note that without this change of variable, it would be considerably more difficult to verify a Lyapunov function for the platoon over the range of possible $\tau_i$ parameters due to the coupling between the state variables and the dynamics parameters. Based on~\cref{eq:new_dynamics} and the definition of the error state $x_i$, it is not hard to show that the error dynamics of the entire platoon is given by
\begin{equation}
    \label{eq:platoon_dynamics}
    x(k+1) = \bar A x(k) + \bar B u(k)
\end{equation}
where $x \in \R^{2N}$ and $u \in \R^N$ are the concatenated error and input vectors, respectively, and 
\begin{equation*}
    \bar A = \bmat{A \\ & \ddots \\ & & A} ,\ \bar B = \bmat{-B \\ B & -B \\ & \ddots & \ddots \\ & & B & -B}.
\end{equation*}
Given~\cref{eq:platoon_dynamics}, it is tempting to use optimal control techniques, \eg, a linear quadratic regulator (LQR). However, these techniques rely on each vehicle having complete knowledge of the control actions $u$ for the entire platoon, which is not practical for a distributed, real-time implementation.

Alternatively, the dynamics given by~\cref{eq:platoon_dynamics} are equivalent to a string of double integrators~\cite{hao2013stability}, thus enabling the use of various insights from the relevant literature to guide control design. For instance, in~\cite{hao2013stability} it was shown that a nonlinear PF controller that saturates performs better than a linear controller. Thus, we can be confident that placing bounds on the control input (which would always be required in practice) can lead to better performance. Furthermore, this motivates the use of learned neural network controllers which can be more expressive than classical nonlinear controllers.

\subsection{Controller learning and verification}

Our goal is to learn a neural network controller $\pi: \R^2 \to \R$ and Lyapunov function $V: \R^{2N} \to \R$ such that the platoon is exponentially stable within some compact region $\calX$ containing the origin. Note that we can bound the control inputs by appending extra layers to the output layer of the neural network (exact formulation is provided in~\cref{app:bounding_neural_network}). From Lyapunov theory, exponential stability within a compact region $\calX$ is guaranteed if
\begin{subequations}
\begin{align}
    & V(0) = 0, \label{eq:lyap_1a} \\
    & V(x) > 0 \quad \forall x \in \calX, \label{eq:lyap_1b} \\
    & V(\bar A x + \bar B \pi(x)) - V(x) \leq -\eps_2 V(x) \quad \forall x \in \calX \label{eq:lyap_1c}
\end{align}
\end{subequations}
where $\pi(x) = (\pi(x_1), \ldots, \pi(x_N))$. Note that depending on how $\calX$ is crafted, we can only guarantee stability within some compact sublevel set of $\calX$.

Condition~(\ref{eq:lyap_1a}) is trivially satisfied by setting the bias terms of the neural network $V$ to zero. For neural network verification, we cannot have strict inequalities in the conditions we would like to verify. Thus we convert the condition~(\ref{eq:lyap_1b}) to the condition $V(x) \geq \eps_1 \norm{x}_1$. Now suppose we have learned a controller $\pi$ and a Lyapunov function $V$ and we would like to verify the conditions~(\ref{eq:lyap_1b}) and~(\ref{eq:lyap_1c}) are true. This can be achieved by solving the optimization problems
\begin{subequations}
\begin{align}
    & \max_{x \in \calX} \calL_\text{pos}(x) = \eps_1 \norm{x}_1 - V(x) \label{eq:lyap_pos_ver} \\
    & \max_{x \in \calX} \calL_\text{dec}(x) = V(\bar A x + \bar B \pi(x)) - (1 - \eps_2) V(x) \label{eq:lyap_dec_ver}
\end{align}
\end{subequations}
and verifying their solutions are equal to zero. Note that the objective functions $\calL_\text{pos}$ and $\calL_\text{dec}$ measure the violations of the positivity Lyapunov condition~(\ref{eq:lyap_1b}) and decreasing Lyapunov condition~(\ref{eq:lyap_1c}), respectively. These two optimization problems can be encoded as mixed-integer linear programs (MILPs) by using mixed-integer formulations of the $\ell_1$-norm and the activation functions within the Lyapunov function and control policy (see~\cref{app:mixed_int_l1,app:mixed_int_leaky_relu}). Thus, the optimization problems given by~\cref{eq:lyap_pos_ver,eq:lyap_dec_ver} can be solved to global optimality. We do this using CVXPY~\cite{diamond2016cvxpy} and Gurobi~\cite{gurobi}.

Next, we need an algorithm to train $V$ and $\pi$ to achieve these objectives. A few different algorithms have been proposed, and they generally revolve around minimizing the loss function given by 
\begin{equation}
    \label{eq:lyapunov_loss}
    \calL_V(x) = \lambda_1 \max(\calL_\text{pos}(x), 0) + \lambda_2 \max(\calL_\text{dec}(x), 0)
\end{equation}
and periodically checking if the solutions to~\cref{eq:lyap_pos_ver,eq:lyap_dec_ver} are zero. Achieving this will ensure the system is exponentially stable when using the learned controller. However, since we are controlling vehicles, we are interested in more than just stability. For example, it is important to minimize the risk of collisions (priority number 1) and ensure passenger comfort (priority number two). Thus, on top of attempting to minimize $\calL_V$, we also penalize values that violate a second loss function $\calL_\pi$ given by 
\begin{equation}
    \label{eq:control_loss}
    \calL_\pi(x(t)) = \sum_{k=0}^H (\calL_\text{safe}(x_k) + \calL_\text{comf}(x_k) + \calL_\text{stab}(x_k))
\end{equation}
which attempts to guide the controller toward some desired behavior. In~\cref{eq:control_loss}, $x(t)$ contains the state/error information for each vehicle at timestep $t$, and we forward simulate the system's dynamics over some time horizon $H$ using $x_0 = x(k)$ (the current state) and the platoon forward dynamics given by~\cref{eq:platoon_dynamics} with $u(k) = \pi(x_k)$. The functions $\calL_\text{safe}$, $\calL_\text{comf}$, and $\calL_\text{stab}$ are hand-designed cost functions that guide the controller towards generating trajectories that are safe, comfortable, and stable, respectively. Though we assume linear dynamics in this paper, this loss function is easily extended to handle general differentiable dynamics functions, including neural network dynamics.

The method we propose to concurrently learn a Lyapunov function and a controller is summarized in algorithm~\ref{alg:guided_alg}. In this algorithm, we simulate the system starting from some error state $x$ until the system reaches the origin. The system has two loops that it iterates through until convergence. The outer loop generates an initial error state for the system by solving either~\cref{eq:lyap_pos_ver} or~\cref{eq:lyap_dec_ver} (or randomly selecting a starting point in $\calX$). The inner loop simulates the system using the controller $\pi$ starting from the initial error state and maintains a dataset of points that violate the Lyapunov conditions. 

Once the inner loop converges, the Lyapunov conditions are checked. If satisfied, we have a learned controller that is certifiably stable for the system. If not, the process repeats. Optionally, we can train on the dataset $\calD$ using stochastic gradient descent (or any other optimizer) to further update $V$ and $\pi$. We have found that augmenting the dataset with random points within $\calX$ can help speed up convergence. Another option, as described in~\cite{dai2021lyapstablenn}, is to augment $\calD$ with any points that violate the Lyapunov conditions during the branching process of the MILP solver. Overall, algorithm~\ref{alg:guided_alg} is very flexible and easily adjusted to incorporate ideas from other works.

\vspace{-4.0mm}
\RestyleAlgo{ruled}
\LinesNumbered
\SetKwComment{Comment}{}{}
\begin{algorithm}
\caption{Guided learning of a controller and Lyapunov function.}
\KwData{Initial $V$, $\pi$, dynamics $f$, $\calD = \emptyset$, $\calX$, simulation environment $\texttt{env}$, $\eps$}\label{alg:guided_alg}
\While{not converged} {
    $x \gets \argmax_{x \in \calX} \{L_\text{pos}(x) \text{ or } L_\text{dec}(x) \}$ \\
    \While{$x \geq \eps$} {
        $l \gets \calL_V(x) + \calL_\pi(x)$ \\
        $\calD \gets \calD \cup \{x\}$ if $\calL_V(x) > 0$ \\
        $V,\, \pi \gets l\texttt{.backwardstep()}$ \\
        $x \gets f(x, \pi(x))$
    }
    $l_1,\, l_2 \gets$ solutions to~\cref{eq:lyap_pos_ver,eq:lyap_dec_ver} \\
    break if $l_1 = 0$ and $l_2 = 0$ \\
    $V,\, \pi \gets \texttt{train}(\calD)$ \hfill \Comment{(optional)}
}
\end{algorithm}
\vspace{-5.0mm}

\section{Experiments}

\subsection{Controller learning}

\begin{figure}
    \centering
    \vspace{-4.0mm}
    \includegraphics[width=0.65\textwidth]{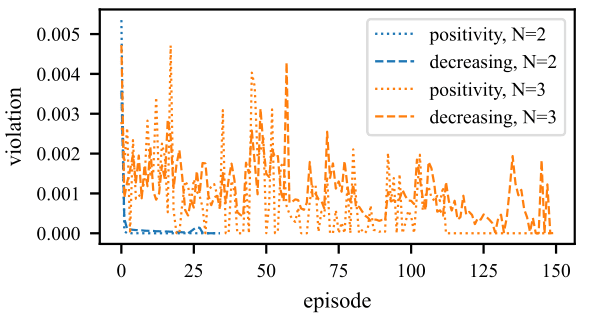}
    \vspace{-2.0mm}
    \caption{Per episode results for the maximum Lyapunov condition violations given by~\cref{eq:lyap_pos_ver} and~\cref{eq:lyap_dec_ver} when running algorithm~\ref{alg:guided_alg}. Results are shown only for $N = 2$ (convergence at episode 30) and $N = 3$ (convergence at episode 150); convergence occurred after 8 episodes for $N = 1$.}
    \label{fig:verification_results}
    \vspace{-4.0mm}
\end{figure}

Algorithm~\ref{alg:guided_alg} returns a feedback controller $\pi: \R^2 \to \R$. For each vehicle $i$, the controller $\pi$ takes in the vehicle's current error state $x_i$, given by~\cref{eq:error_state}, and outputs the vehicle's desired acceleration $u_i$, \ie, $u_i = \pi(x_i)$. We are able to adjust the characteristics of this controller based on how we design $\calL_\pi$. However, we were only able to verify a learned controller for up to $N = 3$. \Cref{fig:verification_results} shows the values of the solutions to~\cref{eq:lyap_pos_ver,eq:lyap_dec_ver} for $N = 2$ and $N = 3$ while running algorithm~\ref{alg:guided_alg}. 

For the experiment shown in~\cref{fig:verification_results}, we parameterized $V$ and $\pi$ using feedforward neural networks with two hidden layers, eight neurons in each hidden layer, and leaky ReLU activation functions with a slope of 0.1. In general, we have found that when using algorithm~\ref{alg:guided_alg}, neural networks with leaky ReLU activation functions converge more easily than neural networks with ReLU activation functions. We see that with $N = 2$, we achieve convergence in about 30 episodes whereas with $N = 3$, we achieve convergence in about 150 episodes. Each time both conditions are equal to zero, we increase the size of the error bounds $\calX$ until we reach the error bounds we are interested in verifying. This explains why for both $N = 2$ and $N = 3$, both conditions reach zero multiple times, but each time except the last are actually for some $\tilde \calX \subset \calX$.

After $N = 3$, we were unable able to achieve convergence even when using a very small $\calX = [-0.1, 0.1]^{Nn}$. We allowed the algorithm to time out after 1000 episodes, which took roughly 18 hours on a computer with an Intel Core i7 and an Nvidia GeForce RTX 3080. Most of this time is spent trying to solve the MILP given by~\cref{eq:lyap_dec_ver} or training on the dataset $\calD$ for 300 epochs. Despite the fact that we are unable to verify the controllers for larger $N$, we have found that the learned controller does still perform well, and it does not take very many episodes for the controller to begin performing well (in only a few episodes for small $N$ and after a few tens of episodes for larger $N$). Thus, for larger $N$, we ignore the verification step and use algorithm~\ref{alg:guided_alg} to directly learn a controller. 

It is possible other methods, like the minimax algorithm described in~\cite{dai2021lyapstablenn}, are better suited to finding a Lyapunov function and controller that guarantees stability. However, it is not immediately clear how we could incorporate soft constraints like the slew constraints into this algorithm, and we found that not including the slew constraints leads to overly reactive controllers that would not be comfortable (\eg, the controller selects large accelerations leading to passenger discomfort~\cite{de2023standards}). Regardless, in~\cite{dai2021lyapstablenn}, the largest state size they were able to certify was for a 3D quadrotor model with 12 states (versus 6 states in this work when $N = 3$), and that took 3 days. We are interested in platoons whose state dimension is much larger. Thus, instead of potentially waiting days or weeks to learn a certificate using our algorithm, we quit early when trying to learn a true Lyapunov function and instead evaluate the learned controller.

\subsection{Baseline controllers}

For the hardware and simulation experiments, we compare against a linear feedback and a DMPC controller. The linear feedback controller computes control inputs using 
\begin{equation*}
    u_i^\text{lin}(t) = k^T x_i(t)
\end{equation*}
where $k \in \R^2$ is a vector of gains and $x_i(t)$ is vehicle $i(t)$'s error state at time $t$. We used $k = (1, 2)$ for all experiments as we have previously found these gains work well on the F1Tenth vehicles \cite{shaham2024design}. The DMPC controller assumes each vehicle receives the planned state trajectory of the vehicle in front of it. Then at each timestep $t$, each vehicle solves an optimization problem that minimizes a weighted objective that penalizes deviation from its predecessor's plan, deviation from its own plan calculated at the previous timestep, and the control input size. See~\cite{shaham2024design,shaham2024dmpc} for implementation details and further analysis of the DMPC controller.

\subsection{Hardware experiments}

To validate our algorithm on hardware, we use the F1Tenth platform, modified as described in~\cite{shaham2024design}. We test the vehicles in a roughly 4 m by 8 m oval racetrack, shown in~\cref{fig:convoy_exp}. We model each vehicle's dynamics using~\cref{eq:og_dynamics} with $\tau_i = 0.3$ (found using a least-squares regression on experimental data~\cite{ljung1998system}). In all experiments, the lead vehicle uses the quadratic cost function DMPC controller described in~\cite{shaham2024design} to track a safe velocity based on the current curvature of the course. This leads to faster trajectories during the straight portions of the course and slower trajectories around the curves. Each following vehicle uses the neural network controller learned using algorithm~\ref{alg:guided_alg} with a desired distance of 0.75 m. 

To ensure safety of the learned controller, we use $L_\text{safe}$ to penalize when a vehicle has a distance of less than 0.25 m to its predecessor. To encourage smooth trajectories with $L_\text{slew}$, we penalize large actions and large changes in speed from one timestep to the next. We also add two ReLU layers and a linear layer to the neural network to ensure the controller only outputs commands such that $\abs{u_i} \leq 3\,\text{ m/s}^2$. To guide the controller toward closed-loop stability, we use $\calL_\text{stab}$ to penalize a running sum of the difference in error from the next timestep to the current timestep over the horizon (because we ideally want the errors to decrease). Other choices for $L_\text{stab}$ can try to guide the controller towards improving the convergence rate of the system to the origin. We ran algorithm~\ref{alg:guided_alg} with $N = 5$, where both $V$ and $\pi$ are two-layer feedforward neural networks with 8 neurons in each hidden layer and leaky ReLU activation functions with slope 0.1 for around 100 episodes to learn a controller.

\begin{figure}
    \centering
    \vspace{-4.0mm}
    \includegraphics[width=0.65\textwidth]{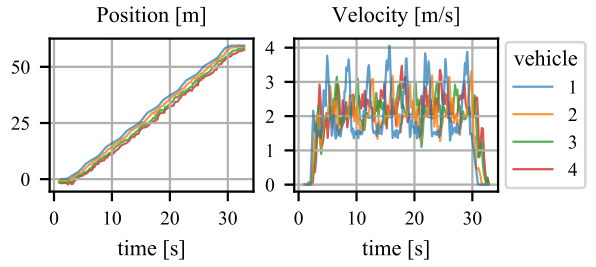}
    \vspace{-2.0mm}
    \caption{Platoon trajectory when using the learned neural network controller.}
    \vspace{-4.0mm}
    \label{fig:platoon_traj_hardware}
\end{figure}

\begin{figure}
    \centering
    \vspace{-4.0mm}
    \includegraphics[width=0.65\textwidth]{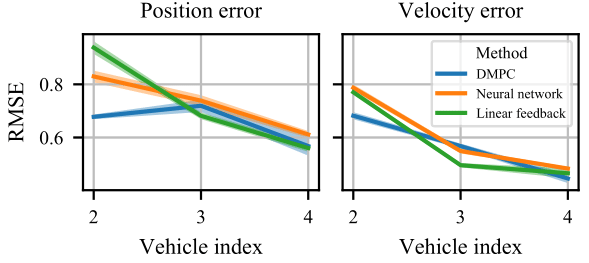}
    \vspace{-2.0mm}
    \caption{Average root-mean-square error for each follower vehicle over the ten trials. We calculate the root-mean-square error for each vehicle's position and velocity error over each trial, and then estimate the root-mean-square error's 95\% confidence interval based on the results over the ten trials.}
    \vspace{-4.0mm}
    \label{fig:rms_error_hardware}
\end{figure}

The platoon's trajectories during a trial run when using the learned controller are shown in~\cref{fig:platoon_traj_hardware}. Since a platoon is perturbed whenever the lead vehicle's velocity changes, we test a scenario where the lead vehicle drives aggressively through the course for 30 seconds before coming to a sudden stop. This results in a trajectory that accelerates to almost 4 m/s during the straight sections of the course and decelerates to around 1.5 m/s while navigating the curves. As we can see, the follower vehicles do not drive as aggressively, since we attempted to guide the controller towards taking aggressive actions if the vehicles are too close (\ie, close to crashing), but didn't penalize being too far away in the same manner. We repeated this experiment ten times for the neural network controller and compared it to the performance for the same quadratic DMPC and linear feedback controllers investigated in~\cite{shaham2024design}. Results comparing the average root-mean-square error (RMSE) for each follower vehicle are shown in~\cref{fig:rms_error_hardware}.

Based on the results in~\cref{fig:platoon_traj_hardware,fig:rms_error_hardware}, it is difficult to determine which algorithm performs the best. We do note that when using the neural network controller, zero collisions occurred between vehicles or with the course boundaries. With the other two controllers, however, vehicles would sometimes collide with the course boundaries. Additionally, with the linear feedback controller, vehicles would sometimes collide with one another when braking at the end of the experiment. We speculate that the collisions with the boundaries occurred because the controllers attempted to follow the predecessor's velocity profile too aggressively when rounding the curves, leading to lateral slipping.

\subsection{Simulation experiments}

For a platoon using a PF topology and a CDH spacing policy, ensuring safety becomes more difficult as the platoon size increases~\cite{seiler2004distprop,zheng2016stabilityhomogeneous,hao2013stability}. To investigate the limits of our approach, we simulate a platoon of $N=100$ vehicles and train a controller using a desired distance of 5 m. We use an almost identical setup as in the hardware section, except we use two hidden layers with 32 neurons each, we penalize a vehicle for getting within 2 m of its predecessor, and we use a heterogeneous platoon where the dynamics parameter $\tau_i \in [0.2, 0.8]$ is selected at random. As mentioned in the previous section, we cannot verify the Lyapunov certificates because 1) the search space becomes very large and 2) the number of binary variables required to solve~\cref{eq:lyap_dec_ver} increases linearly with the platoon size leading to potentially exponential increase in the time required to solve the MILPs. However, we still find that training for a small amount of time (around 100 simulated episodes) yields controllers that perform well in test scenarios. 

As shown in~\cite{shaham2024design}, there is a large gap in the performance of DMPC controllers and linear feedback controllers as platoon size scales. Though DMPC should outperform feedback methods due to the ability to communicate planned trajectories, it would be useful in practice if a feedback controller can perform similarly well. For example, in the scenario where communication is temporarily lost, it would be beneficial to have a feedback controller that can take the place of the DMPC method without disrupting the platoon's motion. Thus, we again compare our learned controller to DMPC and linear feedback, with the goal of bridging the gap between DMPC methods and linear feedback methods.

\begin{figure}
    \centering
    \vspace{-4.0mm}
    \includegraphics[width=0.65\textwidth]{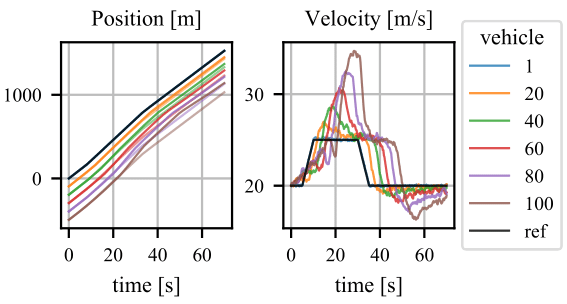}
    \vspace{-2.0mm}
    \caption{Simulated trajectories for a platoon of 100 vehicles when using the learned neural network controller. The lead vehicle tracks the velocity profile shown by the reference. The faded lines in the position plot depict the desired position of each vehicle.}
    \label{fig:platoon_traj_sim}
\end{figure}

For the simulation experiments, the first vehicle tracks a velocity profile that starts at 20 m/s, accelerates to 25 m/s, then decelerates back to 20 m/s. We add Gaussian noise to both the forward dynamics and to the sensing of the position/velocity errors to ensure the learned controller still works in a noisy setting. Note that for a vehicle traveling 20 m/s, common practice would suggest a spacing of 40 m~\cite{vogel2003comparison}, so a 5 m spacing is aggressive for this scenario. The simulation results for the neural network controller are shown in~\cref{fig:platoon_traj_sim}. As expected when using a feedback controller with a PF topology, the errors do propagate down the platoon. However, the controller does stabilize the platoon to the true desired velocities quickly and without any collisions within the platoon. 

\begin{figure}
    \centering
    \vspace{-4.0mm}
    \includegraphics[width=0.65\textwidth]{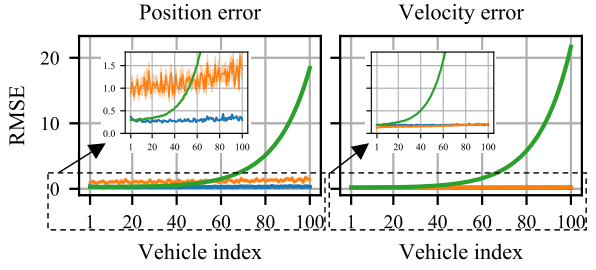}
    \vspace{-2.0mm}
    \caption{RMSE analysis for a simulated platoon of 100 vehicles (blue: DMPC controller, orange: neural network controller, green: linear feedback controller). The zoomed in portion of the graph shows the differences between the feedback and DMPC controllers.}
    \vspace{-4.0mm}
    \label{fig:rms_error_sim}
\end{figure}

Similar to the hardware experiments, we compare the neural network controller to the DMPC controller and a linear feedback controller by simulating the same scenario 10 times and evaluating the RMSE for each vehicle over the 10 trials. We note that collisions did not occur for the DMPC or neural network controllers under this experimental scenario, but the linear feedback controller leads to many collisions for vehicles further down the platoon, and should not be used for large platoons using CDH in practice. Results over the 10 trials are shown in~\cref{fig:rms_error_sim}. Clearly, the neural network controller sacrifices performance for small platoons to obtain improved performance on larger platoons. However, we see that the individual vehicle's spacing RMSE values were higher for smaller platoon sizes and the average values over the 10 trials is much noisier.

\subsection{Discussion}

The algorithm proposed in this work to concurrently learn a neural network controller and a neural network Lyapunov function achieves its goal of generating a controller that significantly outperforms hand-designed controllers. Furthermore, the learned controller translates well from simulation to the real world. However, there is still a need to improve this algorithm (and others proposed in prior works) to achieve convergence for high-dimensional systems, \ie, beyond a state dimension of around ten. From a practical perspective, however, the learned controller typically performs well after only a few tens of episodes, even for very large platoons.

To try to improve the convergence rate of our algorithm, we attempted to ``warm-start'' the Lyapunov function by sharing weights from a previously verified Lyapunov function with fewer vehicles $N$. However, we did not find that this improved performance. Instead, convergence rates were inconsistent even for small $N$. For $N = 2$, the algorithm would typically converge within 50 episodes but could take as many as 200 episodes. For $N = 3$, the algorithm would typically converge within 500 episodes, but could take as many as 1000 episodes. Thus, the actual performance of these algorithms with respect to convergence seems to depend on neural network initialization or the randomly generated dataset. Future research will investigate how to improve the convergence rate.

\section{Conclusion}

We present an algorithm to learn a controller for dynamic systems. When the state dimension is small, the algorithm returns a Lyapunov function that guarantees stability of the closed-loop system. The benefit of our algorithm is the option to design a loss function that guides the controller to behave in a desired way. We apply the algorithm to a platooning problem with an intrinsic tradeoff between traffic performance and safety. Due to a change of variable, we can learn a single controller for a platoon of heterogeneous vehicles, and we investigate the performance of a controller learned using our algorithm on hardware with a platoon of four F1Tenth vehicles and in a simulated environment with 100 vehicles. We show that we can train an algorithm in simulation and obtain performance that bridges the gap between DMPC (the state-of-the-art when vehicles can communicate with one another) and classical linear feedback controllers. 

Future work will consider explicit safety certificates (barrier functions) along with the stability certificates. Furthermore, it is possible that encoding some different notion of state, either by using a history of past errors as input into the neural network or using a recurrent neural network, could lead to better performance. We hope to investigate this using either this algorithm or similar (\eg, reinforcement learning) to investigate the performance limits of feedback controllers for platoons using a CDH spacing policy and a PF topology.


\section*{Acknowledgements}

Research was sponsored by the DEVCOM Analysis Center and was accomplished under Cooperative Agreement Number W911NF-22-2-0001. The views and conclusions contained in this document are those of the authors and should not be interpreted as representing the official policies, either expressed or implied, of the DEVCOM Analysis Center or the U.S. Government. The U.S. Government is authorized to reproduce and distribute reprints for Government purposes notwithstanding any copyright notation herein.


%
%
\bibliographystyle{abbrv}
\bibliography{refs}

\section{Appendix}

\subsection{Bounding the output of a neural network}
\label{app:bounding_neural_network}

Suppose the outputs of a neural network $f: \R^n \to \R^m$ need to be bounded to reside in some box given by a lower bound $l \in \R^m$ and an upper bound $u \in \R^m$. This can be done by applying a clamp function to the output of the neural network, \ie, $y = \textrm{clamp}(f(x); l, u)$ where
\begin{equation*}
    \textrm{clamp}(z; l, u) = \begin{cases}
        l, & z < l \\
        z, & l \leq z \leq u \\
        u, & z > u.
    \end{cases}
\end{equation*}
Since the clamp function can be rewritten in terms of the ReLU activation function,
\begin{equation*}
    \textrm{clamp}(z; l, u) = u - \textrm{ReLU}(u - (\textrm{ReLU}(x - l) + l)),
\end{equation*}
we see that we can bound any neural network to reside in some box by applying two linear + ReLU layers and a third linear layer, all with fixed weights, to the output of the neural network.

\subsection{Mixed-integer formulation of the $\ell_1$-norm}
\label{app:mixed_int_l1}

To convert~\cref{eq:lyap_pos_ver} into a mixed-integer program, we need to convert the $\ell_1$-norm to a mixed-integer form by introducing two variables $t \in \R^n$ and $a \in \{0, 1\}^n$ (where we assume $x \in \R^n$ in~\cref{eq:lyap_pos_ver}) and use the fact that $x$ is assumed to be bounded. Here, we assume $\calX$ is a box with lower bound $l \in \R^n$ and upper bound $u \in \R^n$. Since we require zero is contained in $\calX$, we will always have $l_i \leq 0$ and $u_i \geq 0$ for all $i$. Then the optimization problem $\max_{x \in \calX} \norm{x}_1$ can be equivalently written as
\begin{equation*}
    \begin{array}{ll}
        \underset{t, a}{\textrm{maximize}} & \sum_{i=1}^n t_i \\
        \textrm{subject to} & t \geq x \\
        & t \geq -x \\
        & t \leq x + 2 l \odot (a - 1) \\
        & t \leq 2 u \odot a - x
    \end{array}
\end{equation*}
where $\odot$ denotes the Hadamard product (elementwise multiplication).

\subsection{Mixed-integer formulation of the leaky ReLU}
\label{app:mixed_int_leaky_relu}

The leaky ReLU activation function, denoted by $\sigma$, is given by
\begin{equation*}
    \sigma(z) = \max(\alpha z, z)
\end{equation*}
where $0 \leq \alpha < 1$ is the slope of the function in the left-half plane (note that $\alpha = 0$ corresponds to the ReLU activation). If $z \in \R^n$ then this function acts elementwise on the input. Assuming $z$ is bounded, \ie, $l \leq z \leq u$ for some $l$ and $u$, then by introducing the binary variable $a \in \{0, 1\}^n$, the constraint $y = \sigma(z)$ can be reformulated using the following mixed-integer linear constraints:
\begin{align*}
    y & \geq z \\
    y & \geq \alpha z \\
    y & \leq \alpha z - (\alpha - 1) u \odot a \\
    y & \leq z + (\alpha - 1) l \odot (1 - a).
\end{align*}
If $l_i > 0$ then we must have $a = 1$ and these constraints are equivalent to $y_i = z_i$. Similarly, if $u_i < 0$ then we must have $a = 0$ and these constraints are equivalent to $y_i = \alpha z_i$. The variable $a$ thus indicates if the activation function is in the right-half plane or left-half plane. One way to find bounds $l$ and $u$ on pre-activation variables in a neural network is by using interval bound propagation~\cite{gowal2018effectiveness}, which is the method used in our implementation.

\end{document}